%% file: top.tex
\documentclass[10pt,twocolumn,letterpaper]{article}

\usepackage{iccv}
\usepackage{times}
\usepackage{epsfig}
\usepackage{graphicx}
\usepackage{amsmath}
\usepackage{amssymb}
\usepackage{placeins}
\usepackage{amsmath}

\usepackage{diagbox,tabu,stackengine}
\usepackage[]{algorithm2e}
\usepackage{url}            
\usepackage{amsfonts}       

\usepackage{caption}
\usepackage{multirow}
\usepackage{bbm}

\usepackage[pagebackref=true,breaklinks=true,letterpaper=true,colorlinks,bookmarks=false]{hyperref}
\DeclareMathOperator*{\argmax}{arg\,max}

\usepackage[pagebackref=true,breaklinks=true,letterpaper=true,colorlinks,bookmarks=false]{hyperref}

\newcolumntype{C}[1]{>{\centering\let\newline\\\arraybackslash\hspace{0pt}}m{#1}}

\iccvfinalcopy 
\newcommand{\incircbin}[1]{%
  \mathbin{%
    \mathchoice%
    {\protect\incircint{\displaystyle}{#1}}%
    {\protect\incircint{\textstyle}{#1}}%
    {\protect\incircint{\scriptstyle}{#1}}%
    {\protect\incircint{\scriptscriptstyle}{#1}}%
  }%
}
\newcommand{\incircint}[2]{%
  \ooalign{$#1\bigcirc$\crcr\hidewidth$#1#2$\hidewidth\crcr}%
}
\newcommand{\gatedconv}{\incircbin{*}}

\def\iccvPaperID{3458} 

\ificcvfinal\fi
\begin{document}

\title{Gated-SCNN: Gated Shape CNNs for Semantic Segmentation 
}

\author{
    Towaki Takikawa$^{1,2}\thanks{authors contributed equally}$\hspace{1cm}
    David Acuna$^{1,3,4}\footnotemark[1]$  \hspace{1cm}
    Varun Jampani$^{1}$  \hspace{1cm}
    Sanja Fidler$^{1,3,4}$\\
$^1$NVIDIA \hspace{1em} $^2$University of Waterloo \hspace{1em} $^3$ University of Toronto \hspace{1em}  $^4$Vector Institute  \\
{\tt\small ttakikaw@edu.uwaterloo.ca}, {\tt\small davidj@cs.toronto.edu},
{\tt\small \{vjampani, sfidler\}@nvidia.com}
}

\maketitle

\begin{abstract}
    Current state-of-the-art methods for image segmentation form a dense image representation     
    where the color, shape and texture information are 
    all  processed together inside a deep CNN. This however may not be ideal as they contain very different type of information relevant for recognition. 
Here, we propose a new two-stream CNN architecture for semantic segmentation that explicitly wires shape information as a separate processing branch, \ie \emph{shape stream}, that processes information in parallel to the classical stream.  
Key to this architecture is a new type of gates that connect the intermediate layers of the two streams. Specifically, we use the higher-level activations in the classical stream to gate the lower-level activations in the shape stream, effectively removing noise and helping the shape stream to only focus on processing the relevant boundary-related information. 
This enables us to use a very shallow architecture for the shape stream that operates on the image-level resolution.
Our experiments show that this leads to a highly effective architecture that produces sharper predictions around object boundaries and significantly boosts performance on thinner and smaller objects. 
    Our method achieves state-of-the-art performance on the Cityscapes benchmark, in terms of both mask (mIoU) and boundary (F-score) quality, improving by 2\% and 4\% over strong baselines. \\\small{Project Website: \href{https://nv-tlabs.github.io/GSCNN/}{https://nv-tlabs.github.io/GSCNN/}}

\end{abstract}

\vspace{-2mm}
\input{intro}

\vspace{-0mm}
\input{related}

\input{method}

\input{results}

\input{conc}

\renewcommand{\baselinestretch}{0.8}
\vspace{-1mm}
\selectfont
\begin{footnotesize}
\paragraph{\footnotesize Acknowledgments.}
We thank Karan Sapra for sharing their DeepLabV3+ implementation.

\end{footnotesize}
\renewcommand{\baselinestretch}{1.0}

 \FloatBarrier

{\small
\bibliographystyle{ieee}
\bibliography{egbib}
}

\end{document}

%% file: intro.tex
\vspace{-3mm}
\section{Introduction}
\label{sec:intro}

Semantic image segmentation is one of the most widely studied problems in computer vision with applications 
in 
autonomous driving~\cite{pohlen2016full,kitti,ZhangCVPR16}, 
3D reconstruction ~\cite{MalikM89,Lee2009GeometricRF}
 and image generation~\cite{pix2pix2016,wang2018pix2pixHD} to name 
 a few.
In recent years, Convolutional Neural Networks (CNNs) have led to dramatic improvements in accuracy in almost all the major  segmentation  benchmarks.
A standard practice is to adapt an image classification CNN architecture for the task of semantic segmentation by converting fully-connected layers into convolutional layers~\cite{LongCVPR2015}.
However, using classification architectures for dense pixel prediction has several drawbacks~\cite{dilation,LongCVPR2015,pspnet,deeplabv3plus2018}. One eminent drawback is the loss in spatial resolution of the output due to the use of pooling layers. 
This prompted
several works~\cite{dilation,pspnet,gadde2016superpixel,liu2017learning,huang2017densely} to propose specialized CNN modules 
that help restore the spatial resolution of the network output.

We argue here that there is also an inherent inefficacy in the architecture design since color, shape and texture information are 
    all  processed together inside one deep CNN. Note that these likely contain very different amounts of information that are relevant for recognition. For example, one may need to look at the complete and detailed object boundary to get a discriminative encoding of shape~\cite{PolygonPP2018,curvegcn}, while color and texture contain fairly low-level information. This may also provide an insight of why residual~\cite{he15deepresidual}, skip~\cite{he15deepresidual,yu2018deep} or even dense connections~\cite{huang2017densely} lead to the most prominent performance gains. Incorporating additional connectivity helps the different types of information to flow across different scales of network depth. 
    Disentangling these representations by design may, however, lead to a more natural and effective recognition pipeline. 

\begin{figure}[t!]
 
\centering
\includegraphics[width=\linewidth,trim=0 0 0 0,clip]{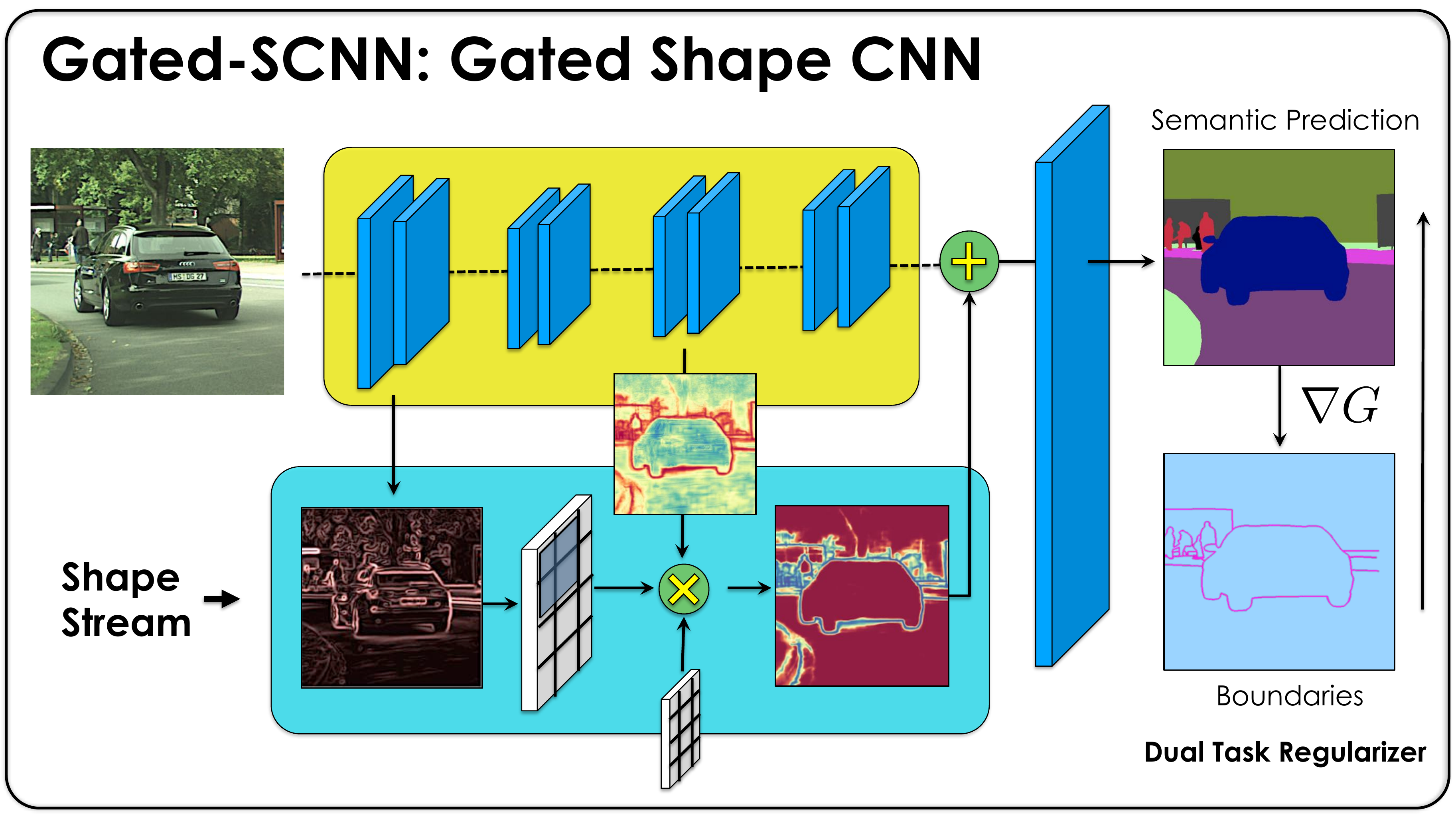} 
\vspace{-6mm}
\caption{We introduce \emph{Gated-SCNN} (GSCNN), a new two-stream CNN architecture for semantic segmentation that explicitly wires shape information as a separate processing stream. 
GSCNN uses a new gating mechanism to connect the intermediate layers.
Fusion of information between streams is done at the very end through a fusion module.
To predict high-quality boundaries, we exploit a new loss function that encourages the predicted
semantic segmentation masks to align with ground-truth  boundaries.}
\label{fig:teaser}
\vspace{-3mm}
\end{figure}

In this work, we propose a new two-stream CNN architecture for semantic segmentation that explicitly wires shape information as a separate processing branch.  In particular, we keep the classical CNN in one stream, and add a so-called \emph{shape stream} that processes information in parallel. We explicitly do not allow fusion of information between the two streams until the very top layers. 

Key to our architecture are a new type of gates that allow the two branches to interact. In particular, we exploit the higher-level information contained in the classical stream to denoise activations in the shape stream in its very early stages of processing. By doing so, the shape stream focuses on processing only the relevant information. This allows the shape stream to adopt a very effective shallow architecture that operates on the full image resolution. To achieve that the shape information gets directed to the desired stream, we supervise it with a semantic boundary loss. 
We further exploit a new loss function that encourages the predicted semantic segmentation to correctly align with the ground-truth semantic boundaries, which further encourages  the fusion layer to exploit information coming from the shape stream. We call our new architecture \emph{GSCNN}. 

We perform extensive evaluation on the Cityscapes benchmark~\cite{cityscapes}. Note that our GSCNN can be used as plug-and-play on top of any classical CNN backbone. In our experiments, we explore ResNet-50 \cite{he15deepresidual}, ResNet-101 \cite{he15deepresidual} and WideResnet \cite{zagoruyko2016wide} and show significant improvements in all. We outperform the  state-of-the-art DeepLab-v3+\cite{deeplabv3plus2018} by more than 1.5 \% in terms of mIoU 
and 4\% in F-boundary score. Our gains are particularly significant for the thinner and smaller objects (\ie poles, traffic light, traffic signs), where we get up to 7\% improvement in terms of IoU. 

We further evaluate performance at varying distances from the camera, using a prior as proxy for distance. Experiments show that we consistently outperform the state-of-the-art baseline achieving up to 6\% improvement in terms of mIoU at the largest distance (\ie further away objects).

%% file: related.tex
\vspace{-1mm}
\section{Related Work}
\label{sec:related}

\vspace{1mm}
\noindent \textbf{Semantic Segmentation.}
State-of-the-art approaches for semantic segmentation are predominantly based on CNNs. 
Earlier approaches~\cite{LongCVPR2015,chen2014semantic} convert classification networks into fully convolutional networks (FCNs) for efficient end-to-end training for semantic segmentation.
Several works~\cite{deeplab,lin2016efficient,zheng2015conditional,SchwingTR2015,HeCVPR2014,Arnab16,liu2015semantic,jampani2016learning,chandra2016fast} propose to use structured prediction modules such as conditional random fields (CRFs) on network output for improving the segmentation performance, especially around object boundaries.
To avoid costly DenseCRF~\cite{krahenbuhl2011efficient}, the work of~\cite{chen2016semantic} uses fast domain transform~\cite{gastal2011domain} filtering on network output while also predicting edge maps from intermediate CNN layers.
We also predict boundary maps to improve segmentation performance. Contrary to~\cite{chen2016semantic}, which uses edge information to refine network output, we inject the learned boundary information into intermediate CNN layers. Moreover, we propose specialized network architecture and a dual-task regularizer to obtain high-quality boundaries.

More recently, dramatic improvements in performance and inference speed have been driven by new architectural designs.
For example, 
PSPNet~\cite{pspnet} and DeepLab \cite{deeplab,deeplabv3plus2018} proposed a feature pyramid pooling module that incorporates multiscale context by aggregating features at multiples scales.
Similar to us,
\cite{pohlen2016full} proposed a two stream network, however, in their case, the main purpose of the second stream
is to recover high-resolution features that are lost with pooling layers. Here, we explicitly specialize the second stream to process shape related information. 
Some works~\cite{gadde2016superpixel,liu2017learning,wang2018non} propose modules that use learned pixel affinities for structured information propagation across intermediate CNN representations. Instead of learning specialized information propagation modules, we propose to learn high-quality shape information through carefully designed network and loss functions. Since we simply concatenate shape information with segmentation CNN features, our approach can be easily incorporated into existing networks for performance improvements.

\vspace{1mm}
\noindent \textbf{Multitask Learning.}
Several works have also explored the idea of 
combining networks for complementary tasks to improve learning efficiency, prediction accuracy and generalization across computer vision tasks.
For example, the works of~\cite{teichmann2018multinet,misra2016cross,kokkinos2017ubernet,kendall2018multi, kong2018recurrent}, proposed unified architectures that learn a shared representation using multi-task losses.
Our main goal is not to train a multi-task network, but to enforce a structured representation that exploits the duality between the segmentation and boundary prediction tasks. 
 \cite{cheng2017fusionnet, bertasius2016semantic} 
 simultaneously learned segmentation and boundary detection network, while~\cite{lin2017refinenet, peng2017large} learned boundaries as an intermediate representation to aid segmentation. Contrary to these works, where semantics and boundary information interact only at the loss functions, we explicitly inject boundary information into segmentation CNN and also propose a dual-task loss function that refines both semantic masks and boundary predictions.
 
\vspace{1mm}
\noindent \textbf{Gated Convolutions.} Recent work on language modeling have also proposed the idea of using 
gating mechanisms in convolutions. For instance, ~\cite{dauphin2017language} proposed to
replace the recurrent connections typically used in recurrent networks with gated temporal convolutions. 
\cite{yu2018free}, on the other hand, proposed the use of convolutions with a soft-gating mechanism for  Free-Form Image Inpainting and \cite{van2016conditional} proposed Gated PixelCNN for conditional image generation.
In our case, we use a gated convolution operator for the task of semantic segmentation and to define the information flow between the shape and regular streams. 

\begin{figure*}[t!]
\vspace{-2mm}
\centering
\includegraphics[width=1\linewidth,trim=0 0 0 0,clip]{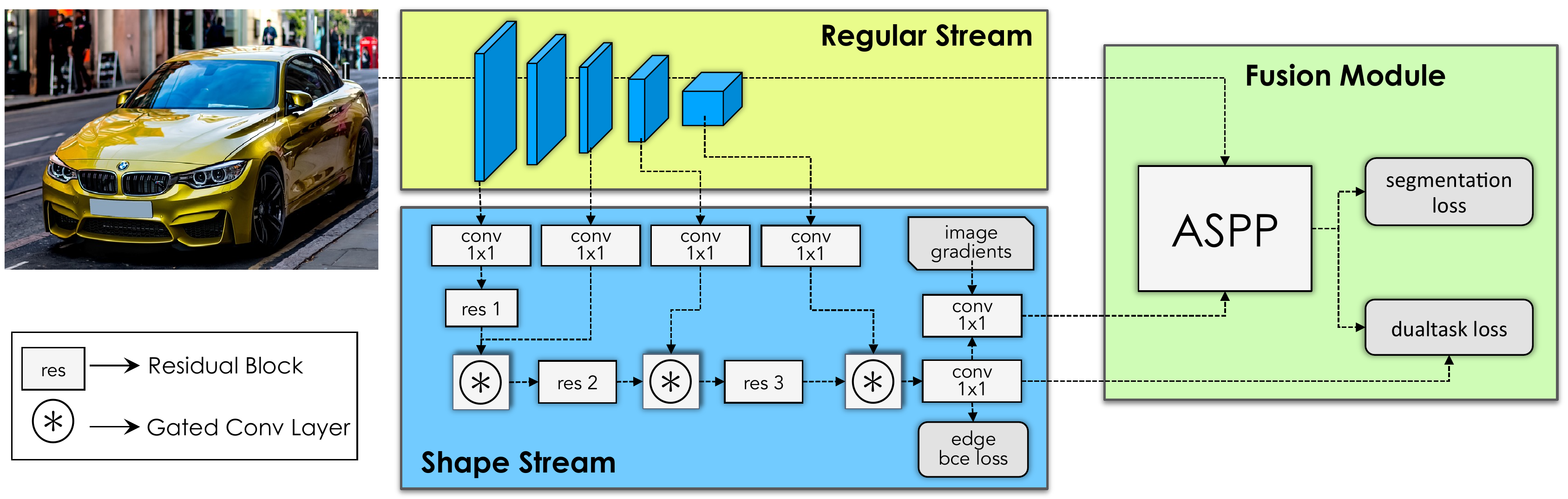}
\vspace{-7mm}
\caption{
	{\bf GSCNN architecture.} Our architecture constitutes of two main streams. The regular stream and the shape stream.
The regular stream can be any backbone architecture. The shape stream focuses on shape processing  through a set of residual blocks, Gated Convolutional Layers (GCL) and supervision. 
A fusion module later combines information from the two streams in a multi-scale fashion using an Atrous Spatial Pyramid Pooling module (ASPP). High quality boundaries on the segmentation masks are ensured through a Dual Task Regularizer . 
 }
\label{fig:nms_layer}
\vspace{-2.5mm}
\end{figure*}

\vspace{-1mm}

%% file: method.tex
\section{Gated Shape CNN}
\label{sec:method}

In this section, we present our Gated-Shape CNN architecture for semantic segmentation. 
As depicted in Fig.~\ref{fig:nms_layer}, our network consists of two streams of networks followed by a fusion module. 
The first stream of the network (``regular stream'') is a standard segmentation CNN, and the second stream (``shape stream'') processes shape information in the form of semantic boundaries. 
We enforce shape stream to only process boundary-related information by our carefully designed Gated Convolution Layer (GCL) and local supervision.
We then fuse semantic-region features from the regular stream and boundary features from the shape stream to produce a refined segmentation result, especially around boundaries.
Next, we describe, in detail, each of the modules in 
our framework 
followed by our novel GCL.

\vspace{1mm}
\noindent \textbf{Regular Stream.} This stream,
denoted as $\mathcal{R}_\theta(I)$, with parameters $\theta$, takes image $I \in \mathbb{R}^{3 \times H \times W}$ with height $H$ and width $W$ as input and produces dense pixel features. The regular stream can be any feedforward fully-convoutional network such as ResNet~\cite{he15deepresidual} based or VGG~\cite{vggcnn} based semantic segmentation network. Since ResNets are the recent state-of-the-art for semantic segmentation, we make use of ResNet-like architecture such as ResNet-101~\cite{he15deepresidual} and WideResNet~\cite{zagoruyko2016wide} for the regular stream.
We denote the output feature representation of the regular stream as $r \in \mathbb{R}^{C \times \frac{H}{m} \times \frac{W}{m}}$ where $m$ is the stride of the regular stream.

\vspace{1mm}
\noindent \textbf{Shape Stream.} 
This stream,
denoted as $\mathcal{S}_\phi$, with parameters $\phi$, takes image gradients $\nabla I$ as well as output of the first convolutional layer of the regular stream as input and produces semantic boundaries as output.
The network architecture is composed of a few residual blocks interleaved with gated convolution layers (GCL). GCL, explained below, ensures that the shape stream only processes boundary-relevant information.
We denote the output boundary map of the shape stream as $s \in \mathbb{R}^{H \times W}$.
Since we can obtain ground-truth (GT) binary edges from GT semantic segmentation masks, we use supervised
binary cross entropy loss on output boundaries to supervise the shape stream.

\vspace{1mm}
\noindent \textbf{Fusion Module.} This module, denoted as $\mathcal{F}_\gamma$, with parameters $\gamma$, takes as input the dense feature representation $r$ coming from the regular branch and fuses it with the boundary map $s$ output by the shape branch in a way that multi-scale contextual information is preserved. It combines region features with boundary features and outputs a refined semantic segmentation output. More formally, for a segmentation prediction of $K$ semantic classes, it outputs a categorical distribution $f = p(y|s,r)=\mathcal{F}_{{\gamma}}(s,r) \in \mathbb{R}^{K \times H \times W}$, which represents the probability that pixels belong to each of the $K$ classes.
Specifically, we  merge the boundary map $s$ with $r$ using an Atrous Spatial Pyramid Pooling~\cite{deeplabv3plus2018}. 
This allows us to preserve the multi-scale contextual information and is proven to be an essential component in state-of-the-art semantic segmentation networks.

\vspace{-2mm}
\subsection{Gated Convolutional Layer}
Since the tasks of estimating semantic segmentation and semantic boundaries are closely related, we devise a novel GCL layer that facilitates flow of information from the regular stream to the shape stream.
GCL is a core component of our architecture and helps the shape stream to only process relevant information by filtering out the rest. 
Note that the shape stream does not incorporate features from the regular stream. Rather, it uses GCL to deactivate its own activations that are not deemed relevant by the higher-level information contained in the regular stream. One can think of this as a collaboration between two streams, where the more powerful one, which has formed a higher-level semantic understanding of the scene, helps the other stream to focus only on the relevant parts since start. This enables the shape stream to adopt an effective shallow architecture that processes the image at a very high resolution.

We use GCL in multiple locations between the two streams.  Let $m$ denote the number of locations, and 
let $t \in {0,1,\cdots,m}$ be a running index where $r_t$ and $s_t$ denote intermediate representations of the corresponding regular and shape streams that we process using a GCL.
To apply GCL, we first obtain an attention map $\alpha_t \in \mathbb{R}^{H\times W}$ by concatenating $r_t$ and $s_t$ followed by a normalized $1\times1$ convolutional layer $C_{1\times1}$ which in turn is followed by a sigmoid function $\sigma$ :
\vspace{-2mm}
\begin{align}
    \alpha_t &= \sigma(C_{1\times1}(s_t || r_t)), \label{eqn:attention}
\end{align}
where $||$ denotes concatenation of feature maps.
Given the attention map $\alpha_t$, GCL is applied on $s_t$ as an element-wise product $\odot$ with attention map $\alpha$ followed by a residual connection and channel-wise weighting with kernel $w_t$. At each pixel $(i,j)$, GCL $\gatedconv$ is computed as
\vspace{-2mm}
\begin{align}
\begin{aligned}
\hat{s}_{t}^{(i,j)}& = {(s_t \gatedconv w_t)_{(i,j)}}  \\
	           & = ((s_{t_{(i,j)}} \odot \alpha_{t_{(i,j)}}) + s_{t_{(i,j)}})^T w_t.
\label{eqn:gated_conv}
\end{aligned}
\end{align}
$\hat{s}_t$ is then passed on to the next layer in the shape stream for further processing.
Note that both the attention map computation and gated convolution are differentiable and therefore backpropagation can be performed end-to-end. 
Intuitively, $\alpha$ can also be seen as an attention map that weights more heavily areas with important boundary information. 
In our experiments, we  use three GCLs and connect them to the third, fourth and last layer of the regular stream. Bilinear interpolation, if needed,  is used to upsample the feature maps coming from the regular stream.

\subsection{Joint Multi-Task Learning}
We jointly learn the regular and shape streams together with the fusion module in an end-to-end fashion. 
We jointly supervise segmentation and boundary map prediction during training. 
Here, the boundary map is a binary representation of all the outlines of objects and stuff classes in the scene (Fig~\ref{fig:boundary}). 
We use standard binary cross-entropy (BCE) loss on predicted boundary maps $s$ and use standard cross-entropy (CE) loss on predicted semantic segmentation $f$:
\begin{align}
\label{eqn_multi_task_loss}
\mathcal{L}^{\theta\,\phi,\gamma}= \lambda_1 \mathcal{L}^{\theta,\phi}_{BCE}(s, \hat{s})+\lambda_2 \mathcal{L}^{\theta\,\phi,\gamma}_{CE}(\hat{y},f)
\end{align}
where $\hat{s} \in \mathbb{R}^{H \times W}$ denotes GT boundaries and $\hat{y} \in \mathbb{R}^{H \times W}$ denotes GT semantic labels.
Here, $\lambda_1,\lambda_2$ are two hyper-parameters that control the weighting between the losses.

As depicted in Fig.~\ref{fig:nms_layer}, the BCE supervision on boundary maps $s$ is performed before feeding them into the fusion module. Thus the BCE loss $\mathcal{L}^{\theta,\phi}_{BCE}$ updates the parameters of both the regular and shape streams.
The final categorical distribution $f$ of semantic classes is supervised with CE loss $\mathcal{L}^{\theta\,\phi,\gamma}_{CE}$ at the end as in standard semantic segmentation networks, updating all the network parameters.
In the case of $BCE$ on boundaries, we follow \cite{xie2015hed,yu2017casenet} and use a coefficient $\beta$ to account for the high imbalance between boundary/non-boundary pixels. 

\subsection{Dual Task Regularizer} 
As mentioned above, $p(y|r,s) \in R^{K\times H\times W}$ denotes a categorical distribution output of the fusion module.
Let $\zeta \in R^{H\times W}$ be a potential that represents whether a particular pixel belongs to a semantic boundary in the input image $I$. It is computed by taking a spatial derivative on segmentation output as follows:

\vspace{-5mm}
\begin{align}
	\label{eqn:g}
\zeta &=\frac{1}{\sqrt{2}}||\nabla(G*\argmax_k p(y^k| r,s))||
\end{align}
where $G$ denotes Gaussian filter. If we assume $\hat \zeta$ is a GT binary mask computed in the same way from the GT semantic labels $\hat{f}$, we can write the following loss function:

\vspace{-3mm}
\begin{align}
\label{eqn:loss_reg_edges_domain}
\mathcal{L}^{\theta\,\phi,\gamma}_{reg_\rightarrow}=\lambda_3 \sum_{p^+} |\zeta(p^+ ) -\hat \zeta(p^+)|
\vspace{-3mm}
\end{align} 
where $p^{+}$ 
contains the set of all non-zero pixel coordinates in both $\zeta$ and $\hat \zeta$.
Intuitively, we want to ensure that boundary pixels are penalized when there is a mismatch with GT boundaries,
and to avoid non-boundary pixels to dominate the loss function. 
Note that the above regularization loss function exploits the duality between boundary prediction and semantic segmentation in the boundary space.

Similarly, we can use the boundary prediction from the shape stream $s \in \mathbb{R}^{H\times W}$ to ensure consistency between the binary boundary  prediction $s$ and the predicted semantics $p(y|r,s)$:
\vspace{-3mm}
\begin{align}
\label{eqn:loss_reg_semantic_domain}
\mathcal{L}^{\theta\,\phi,\gamma}_{reg_\leftarrow} &=\lambda_4  \sum_{k,p} \mathbbm{1}_{s_p} [\hat{y}^k_{p}\log p(y^k_p| r,s) ],
\vspace{-3mm}
\end{align} 
where $p$ and $k$ runs over all image pixels and semantic classes, respectively. $ \mathbbm{1}_{s} = \left\{ \begin{array}{@{}l@{\thinspace}l} 1: s > \text{\emph{thrs}}  \end{array} \right \}$  corresponds to the indicator function and \emph{thrs} is a confidence threshold, we use 0.8 in our experiments. The total dual task regularizer loss function can be written as:
\begin{align}
\label{eqn:loss_reg}
\mathcal{L}^{\theta\,\phi,\gamma}= \mathcal{L}^{\theta\,\phi,\gamma}_{reg_\rightarrow}  +\mathcal{L}^{\theta\,\phi,\gamma}_{reg_\leftarrow} 
\end{align} 
Here, $\lambda_3$ and $\lambda_4$ are two hyper-parameters that control the weighting of the regularizer.

\begin{figure*}[t!]
\vspace{-0mm}
\centering
\begin{minipage}{0.3\linewidth}
\includegraphics[width=1\linewidth,trim=0 0 0 0,clip]{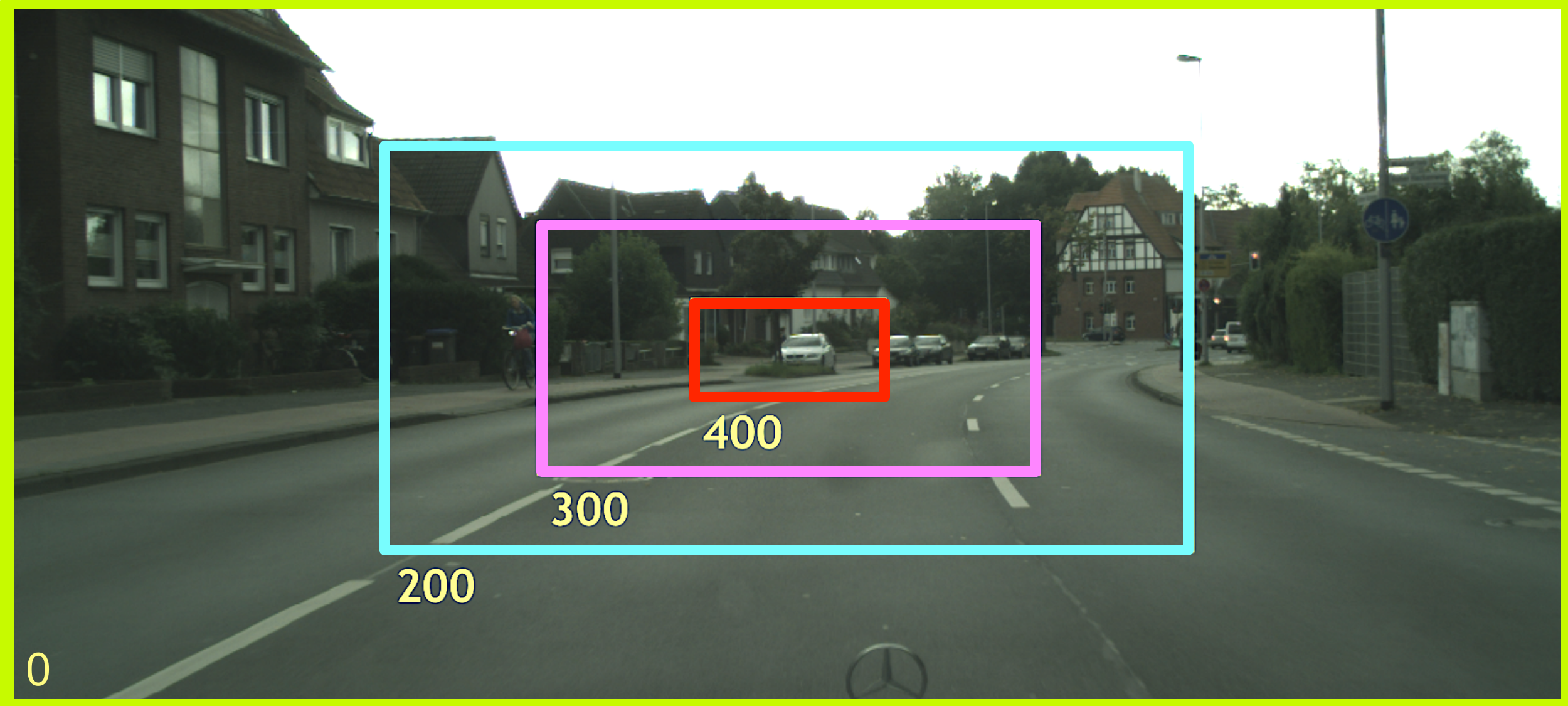}
\vspace{-7mm}
\caption{\small Illustration of the crops used for the distance-based evaluation. 
}
\label{fig:cropping}
\end{minipage}
\hspace{0.2mm}
\begin{minipage}{0.3\linewidth}
\includegraphics[width=\linewidth,trim=0 0 0 0,clip]{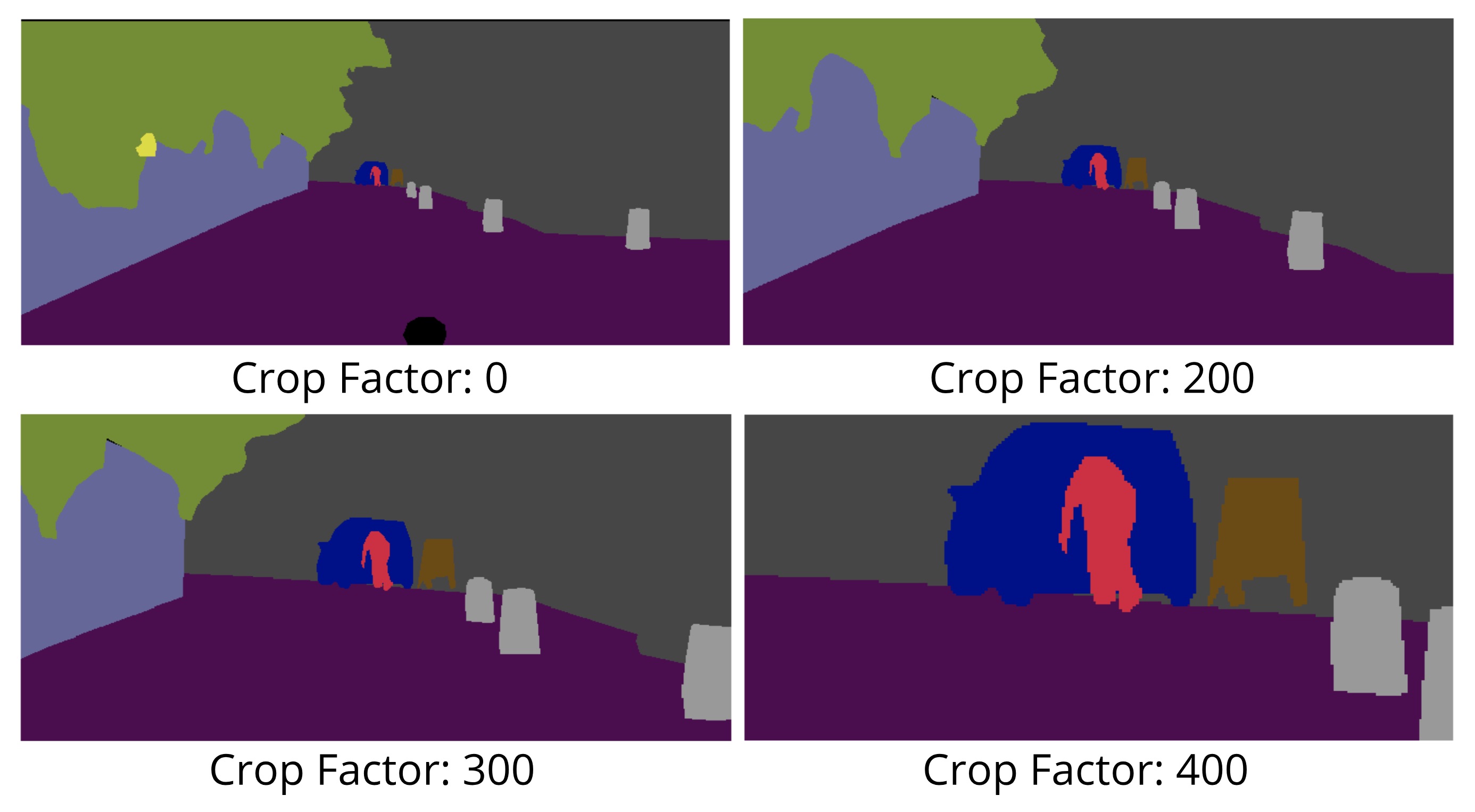}
\vspace{-7.5mm}
\caption{\small Predictions at diff. crop factors.  
}
\label{fig:crop}
\end{minipage}
\hspace{0.2mm}
\begin{minipage}{0.378\linewidth}
\includegraphics[width=\linewidth,trim=20 90 0 20,clip]{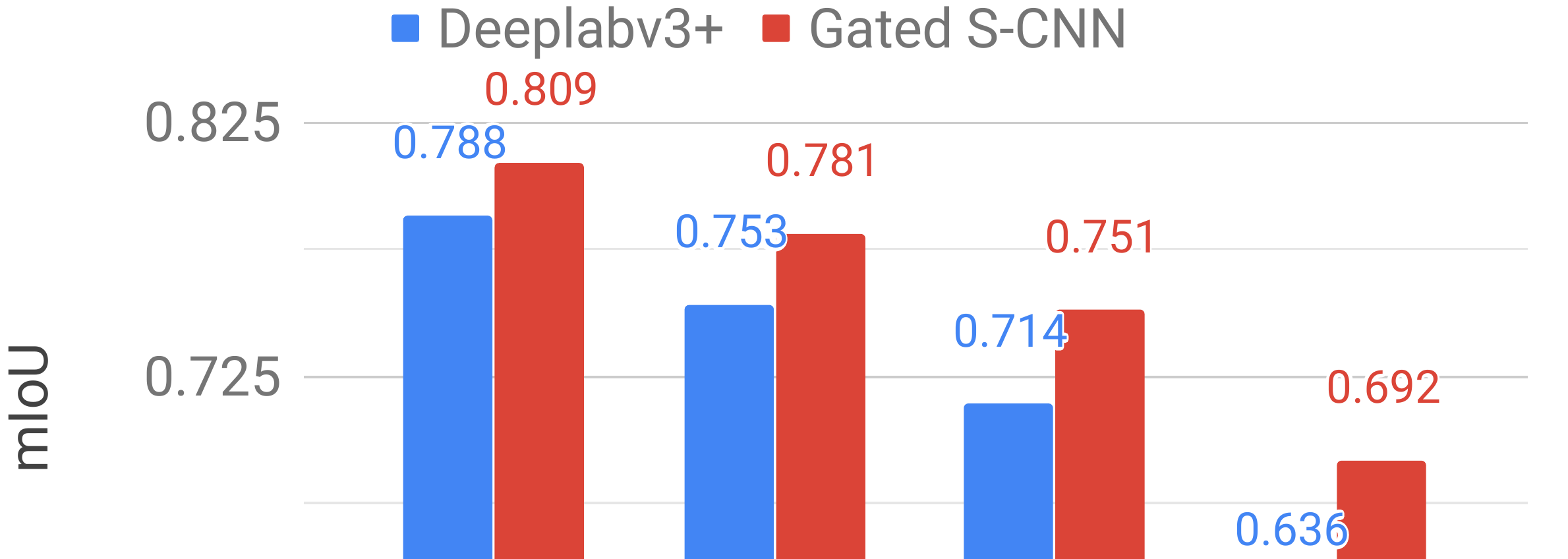}
\vspace{-6.5mm}
\caption{\small {\bf Distance-based evaluation}: Comparison of mIoU at different crop factors.
}
\label{fig:crops}
\end{minipage}
\vspace{-1mm}
\end{figure*}

\begin{table*}[t!]
\centering
\addtolength{\tabcolsep}{-2pt}
\resizebox{\textwidth}{!}{\begin{tabular}{l| c|c|c|c|c|c|c|c|c|c|c|c|c|c|c|c|c|c|c|c}
  Method &   road & s.walk & build. & wall & fence & pole & t-light & t-sign & veg & terrain & sky & person & rider & car & truck & bus & train & motor & bike & mean \\
\hline \hline
        LRR \cite{ghiasi2016laplacian} & 97.7 &79.9& 90.7 &44.4 &48.6 &58.6& 68.2& 72.0& 92.5& 69.3& 94.7& 81.6 &60.0 &94.0& 43.6& 56.8& 47.2 &54.8& 69.7& 69.7 \\
        
        DeepLabV2 \cite{deeplab} & 97.9& 81.3& 90.3& 48.8 &47.4 &49.6& 57.9& 67.3& 91.9& 69.4& 94.2& 79.8 &59.8& 93.7& 56.5& 67.5& 57.5& 57.7& 68.8& 70.4 \\

        Piecewise \cite{lin2016efficient} & 98.0& 82.6& 90.6 &44.0 &50.7 &51.1& 65.0& 71.7& 92.0& 72.0& 94.1& 81.5& 61.1& 94.3 &61.1 &65.1& 53.8& 61.6& 70.6& 71.6 \\

        PSP-Net \cite{pspnet} & 98.2 & 85.8 & 92.8 & 57.5 & 65.9 & 62.6 & 71.8 & 80.7 & 92.4 & 64.5 & 94.8 & 82.1 & 61.5 & 95.1 & 78.6 & 88.3 & 77.9 & 68.1 & 78.0 & 78.8  \\ 
         DeepLabV3+ \cite{deeplabv3plus2018}  & 98.2  &  84.9  & 92.7& \bf{57.3} & 62.1 &  65.2  & 68.6 & 78.9 & 92.7 & 63.5 & 95.3 & 82.3 & 62.8 & 95.4 & \bf{85.3} & 89.1 & 80.9 & 64.6& 77.3 & 78.8   \\
\hline
  \textbf{Ours} (GSCNN)&  \bf{98.3}   & \bf{86.3} & \bf{93.3} & {55.8} & \bf{64.0} & \bf{70.8} & \bf{75.9} &  \bf{83.1} & \bf{93.0} & \bf{65.1}& \bf{95.2} &  \bf{85.3} &  \bf{67.9}& \bf{96.0} &  80.8  &  \bf{91.2} & \bf{83.3} & \bf{69.6} & \bf{80.4}  & \bf{80.8}\\

\hline \hline

\end{tabular}
}
\vspace{-3mm}
\caption{ Comparison in terms of IoU  vs state-of-the-art baselines on the Cityscapes val set.}
 \label{cityscapes_sota_val_set_iou}
 \vspace{-1mm}
\end{table*}

\begin{table*}[t!]
\centering
\addtolength{\tabcolsep}{-2pt}
\resizebox{\textwidth}{!}{\begin{tabular}{c|l| c|c|c|c|c|c|c|c|c|c|c|c|c|c|c|c|c|c|c|c}
Thrs & Method &   road & s.walk & build. & wall & fence & pole & t-light & t-sign & veg & terrain & sky & person & rider & car & truck & bus & train & motor & bike & mean \\
\hline \hline

\multirow{3}{0.1\linewidth}{\centering{12px}}

&        DeepLabV3+ & \bf{92.3} & 80.4 & 87.2 & 59.6 & 53.7 & 83.8 &
75.2 & 81.2 & 90.2 & 60.8 & 90.4 & 76.6 &
78.7 & 91.6 & \bf{81.0} & 87.1 & 92.6 & \bf{81.8} & 78.0 & 80.1
 \\  
&        Ours & 92.2 & \bf{81.7} & \bf{87.9} & \bf{59.6} & \bf{54.3} & \bf{87.1} &
\bf{82.3} & \bf{84.4} & \bf{90.9} & \bf{61.1} & \bf{91.9} & \bf{80.4} &
\bf{82.8} & \bf{92.6} & 78.5 & \bf{90.0} & \bf{94.6} & 79.1 & \bf{82.2} & \bf{81.8}\\

\hline \hline

\multirow{3}{0.1\linewidth}{\centering{9px}}

&        DeepLabV3+ & 91.2 & 78.3 & 84.8 & 58.1 & 52.4 & 82.1 &
73.7 & 79.5 & 87.9 & 59.4 & 89.5 & 74.7 &
76.8 & 90.0 & \bf{80.5} & 86.6 & 92.5 & \bf{81.0} & 75.4 & 78.7 \\  
&        Ours & \bf{91.3} & \bf{80.1} & \bf{86.0} & \bf{58.5} & \bf{52.9} & \bf{86.1} &
\bf{81.5} & \bf{83.3} & \bf{89.0} & \bf{59.8} & \bf{91.1} & \bf{79.1} &
\bf{81.5} & \bf{91.5} & 78.1 & \bf{89.7} & \bf{94.4} & 78.5 & \bf{80.4} & \bf{80.7}  \\  

\hline \hline

\multirow{3}{0.1\linewidth}{\centering{5px}}
&        DeepLabV3+ & 88.1 & 72.6 & 78.1 & 55.0 & 49.1 & 77.9 &
69.0 & 74.7 & 81.0 & 55.8 & 86.4 & 69.0 &
71.9 & 85.4 & \bf{79.4} & 85.4 & 92.1 & \bf{79.4} & 68.4 & 74.7 \\  
&        Ours & \bf{88.7} & \bf{75.3} & \bf{80.9} & \bf{55.9} & \bf{49.9} &\bf{83.6} &
\bf{78.6} & \bf{80.4} & \bf{83.4} & \bf{56.6} & \bf{88.4} & \bf{75.4} &
\bf{77.8} & \bf{88.3} & 77.0 & \bf{88.9} & \bf{94.2} & 76.9 & \bf{75.1} & \bf{77.6}  \\

\hline \hline

\multirow{3}{0.1\linewidth}{\centering{3px}}
&        DeepLabV3+ & 83.7 & 65.1 & 69.7 & 52.2 & 46.2 & 72.0 &
62.8 & 67.7 & 71.8 & 52.0 & 80.9 & 61.5 &
66.4 & 78.8 & \bf{78.2} & 83.9 & 91.7 & \bf{77.9} & 60.9 & 69.7 \\  
&        Ours & \bf{85.0} & \bf{68.8} & \bf{74.1} & \bf{53.3} & \bf{47.0} & \bf{79.6} &
\bf{74.3} & \bf{76.2} & \bf{75.3} & \bf{53.1} & \bf{83.5} & \bf{69.8} & \bf{73.1} & \bf{83.4} & 75.8 & \bf{88.0} & \bf{93.9} & 75.1 & \bf{68.5} & \bf{73.6}   \\  
\hline \hline

\end{tabular}
}
\vspace{-3mm}
\caption{ Comparison vs baselines at different thresholds in terms of boundary F-score  on the 
 Cityscapes val set. \label{cityscapes_sota_val_set_fscore}}
 \vspace{-2mm}
\end{table*}

\vspace{-2mm}
\subsubsection{Gradient Propagation during Training}
In order to back-propagate through Eq~\ref{eqn:loss_reg}, we need to compute the gradients of Eq~\ref{eqn:g}. Letting $g=||.||$, the partial derivatives with respect to a given parameter $\eta$ can be  computed as follows:
\begin{align}
\frac{\partial L}{\partial \eta_i}&=\sum_{j,l} \nabla G* \frac{\partial L}{\partial \zeta_j} \frac{\partial \zeta_j }{\partial g_l} \frac{  \partial \argmax_k p(y^{k})_l}{\partial \eta_i   }
\end{align}
Since $\argmax$ is not a differentiable function we use the Gumbel softmax trick~\cite{jang2016categorical}. 
During the backward pass, we approximate the argmax operator with a softmax with temperature $\tau $:
\begin{align}
\frac{ \partial \argmax_k p(y^{k})} {\partial \eta_i   }&=  \nabla_{\eta_i} \frac{ \exp((\log p(y_k) + g_k )/\tau)}{\sum_j \exp((\log p(y_j) + g_j )/\tau) }
\end{align}
where $g_j \sim \textnormal{Gumbel(0,I)} $ and $\tau$ a hyper-parameter. 
The operator $\nabla G *$ can be computed by filtering with Sobel kernel.

%% file: results.tex
\vspace{-1mm}

\section{Experimental Results}

In this section, we provide an extensive evaluation of
each component of our framework on the challenging Cityscapes dataset~\cite{cityscapes}.
We further show the effectiveness of our approach for several backbone architectures and provide 
qualitative results of our method.

 \vspace{-3mm}
\paragraph{Baselines.}
We use DeepLabV3+ \cite{deeplabv3plus2018}, as our main baseline. This 
consitutes the state-of-the-art architecture for semantic segmentation
and
pretrained models are available.
In most of our experiments, we use our own PyTorch implementation of DeeplabV3+ which differs from \cite{deeplabv3plus2018} in the choice of the backbone architecture. 
Specifically, we use ResNet-50, ResNet-101 and WideResNet as the backbone architecture for our version of DeeplabV3+.
For a fair comparison, when applicable, we refer to this as \emph{Baseline} in our tables.
Additionally,  we also compare against 
published state-of the-art-methods on the validation set and in the Cityscapes benchmark (test set). 
\vspace{-3mm}
\paragraph{Dataset.} 
All of our experiments are conducted on the Cityscapes dataset.
This dataset contains images 
from 27 cities in Germany and neighboring countries. It
contains 2975 training, 500 validation and 1525 test images. 
Cityscapes additionally includes 20,000 additional coarse annotations
(i.e., coarse polygons covering individual objects).
Notice that we supervise our shape stream with boundary ground-truth, and thus the coarse subset is not ideal for our setting. We thus do not use coarse data in our experiments. 
The dense pixel annotations include 30 classes which frequently occur in urban street scenes, out of which 19 are used for the actual training and evaluation. 
We follow~\cite{yu2017casenet,yu2018seal,david19edges} to generate the ground truth
boundaries and supervise our shape stream. 
%

\vspace{-4mm}
\paragraph{Evaluation Metrics.}
We use three quantitative measures to evaluate the performance  of our approach. 
\textbf{1)} We use the widely used intersection over union (IoU) to evaluate whether the network accurately predicts regions.
\textbf{2)} Since our method aims to predict high-quality boundaries, we include another metric for evaluation. Specifically, we follow the boundary metric proposed in~\cite{perazzi2016benchmark} to evaluate the quality of our semantics boundaries.
This metric computes the F-score along the boundary of the predicted mask, given a small slack in distance.
In our experiments, we use thresholds 0.00088, 0.001875, 0.00375, and 0.005 which correspond to 3, 5, 9, and 12 pixels respectively. 
Similarly to the IoU calculation, we also remove void areas 
during the computation of the F-score. Since 
boundaries 
are not provided for the test-set, we use the validation set to compute F-Scores as a metric for boundary alignment.
\textbf{3)}  We use distance-based evaluation in terms of IoU, explained below,  in order to evaluate the performance of the segmentation models at varying distances from the camera. 

\begin{table}
\vspace{-3mm}
\begin{minipage}{0.45\textwidth}
\resizebox{\linewidth}{!}{
\begin{tabular}{c| c| c|c|c}
Metric & Method & ResNet-50 &  ResNet-101   &  Wide-ResNet \\
\hline \hline
\multirow{4}{0.07\linewidth}{\centering{mIoU}}  
 & Baseline &  71.3   &  72.7  & 79.2 \\
\cline{2-5}

 & + GCL &  72.9 & 74.3 & 79.8 \\
& + Gradients & \bf{73.0}    &  \bf{74.7} &   \bf{80.1} \\
\hline \hline

\multirow{4}{0.07\linewidth}{\centering{F-Score}}
   & Baseline & 68.5 & 69.8 & 73.0 \\
\cline{2-5}
 & + GCL    &  71.7 & \bf{73.3} & \bf{75.9} \\
 & + Gradients &\bf{71.7}& 73.0 &  75.6 \\
\hline \hline

\end{tabular}
}
\centering
\vspace{-3mm}
\caption{Comparison of the shape stream, GCL, and additional image gradient features (Canny) for different regular streams. Score on Cityscapes val (\%)  represents mean over all classes and F-Score represents boundary alignment (th=5px).}
\label{tbl_effect_gcl_hf_gcl}
\end{minipage}
\end{table}

\begin{table}
\vspace{-2mm}
\begin{minipage}{0.45\textwidth}
\begin{small}
\begin{tabular}{ c|c| c|c|c}
Method & th=3px & th=5px  & th=9px & th=12px  \\
\hline \hline
   Baseline & 64.1 & 69.8  & 74.8   & 76.7  \\
  \hline
   GCL & 65.0 & 70.8  & 75.9  &  77.8  \\
   + Dual Task & \bf{68.0} & \bf{73.0}   &  \bf{77.2} & \bf{78.8} \\
\hline \hline
\end{tabular}
\end{small}
\centering
\vspace{-3mm}
\caption{Effect of the Dual Task Loss at difference thresholds in terms of boundary quality (F-score). ResNet-101 used in regular stream. }
\label{tbl_effect_dual}
\end{minipage}
\vspace{-2mm}
\end{table}

\begin{table}
\vspace{0mm}
\begin{minipage}{0.45\textwidth}
\resizebox{\linewidth}{!}{
\begin{tabular}{ c |c| c|c}
Base Network & Param   $\Delta$ (\%)& Perf $\Delta$ (mIoU) & Perf $\Delta$ (mF)\\
\hline \hline
ResNet-50 & +0.43 & +1.7 & +3.2\\
ResNet-101 & +0.29 & +2.0 & +3.5\\
WideResNet38 & +0.13 & +0.9 & +2.1\\
\hline \hline
\end{tabular}
}
\centering
\vspace{-3mm}
\caption{Performance improvements and the percentage increase in the number of parameters due to the shape stream on different base networks.}
\label{tbl_params}
\end{minipage}
\vspace{-6mm}
\end{table}

\vspace{-4mm}
\paragraph{Distance-based Evaluation.}
We argue that high accuracy is also important for small (distant) objects, where however, the global IoU metric does not well reflect this. 
Thus, we take crops of varying size around an approximate (fixed) vanishing point as a proxy for distance. 
In our case, this is performed by cropping 100 pixels along each image side except for the top, and the center of the resulting crop is our approximate  vanishing point.
Then, given a predefined cropping factor $c$
, crops are applied such that:  we crop $c $ from the top and bottom and $c\times 2$ from the left and right. 
Intuitively,  a smaller centered crop puts a larger weighting on the smaller objects farther away from the camera. 
An illustration of the procedure is shown in Fig~\ref{fig:cropping}. Fig~\ref{fig:crop} shows example predictions in each of the crops, illustrating how the metrics can focus on evaluating object at different sizes. 
 
 \vspace{-3mm}
 \paragraph{Implementation Details.}
In most of our experiments, we follow the methodology of Deeplab v3+ \cite{deeplabv3plus2018} but use simpler encoders as described in the experiments. 
All our networks are implemented in PyTorch. 
We use  $800 \times 800$ as the training resolution and synchronized batch norm. Training is done on an NVIDIA DGX Station using 8 GPUs with a total batch size of 16.
For Cityscapes, we use a learning rate of 1e-2 with a polynomial decay policy. 
We run the training for 100 epochs for the ablation purposes, and showcase our best results in Table~\ref{cityscapes_sota_val_set_iou} at 230 epochs. 
For our joint loss, we set $\lambda_1 = 20$, $\lambda_2 = 1$, $\lambda_3 = 1$ and $\lambda_4 = 1$. We set $\tau = 1$ for the Gumbel softmax.
All our experiments are conducted in the Cityscapes fine set.

\subsection{Quantitative Evaluation}

In Table~\ref{cityscapes_sota_val_set_iou}, we compare the performance of our GSCNN against the baselines in terms of region accuracy (measured by mIoU). The numbers are reported on the validation set, and computed on the full image (no cropping). 
In this metric, we achieve a 2\% improvement, which is a significant result at this level of performance.
In particular, we notice that we obtain significant improvements for small objects:
motorcycles, traffic signs, traffic lights, and poles. 

Table~\ref{cityscapes_sota_val_set_fscore}, on the other hand, compares the performance of our method against the baseline in terms of boundary accuracy (measured by F score). 
Similarly, our model performs considerably better, outperforming the baseline by close to 4\% in the strictest regime. 
Note that, for fair comparison, we only report models trained on the Cityscapes fine set. Inference for all models is done on a single-scale.

  \begin{figure}[t!]
\vspace{-2mm}
\centering
\includegraphics[width=\linewidth,trim=0 0 0 0,clip]{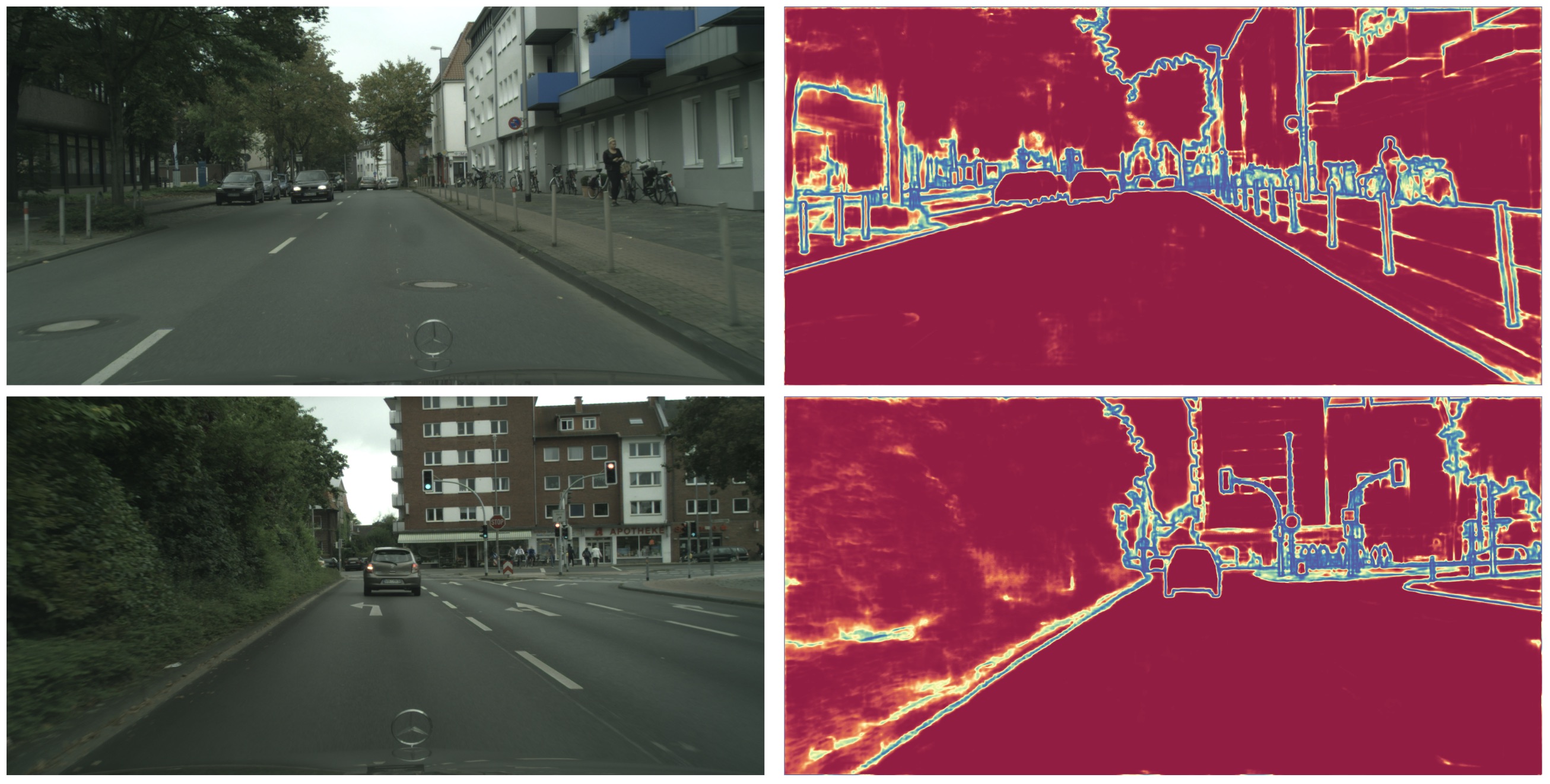}
\vspace{-7mm}
\caption{Example output of shape stream fed into the fusion module.
}
\label{fig:boundary}
\vspace{-2mm}
\end{figure}

\begin{figure*}[t!]
 
\centering
\includegraphics[width=1\linewidth,trim=0 0 0 0,clip]{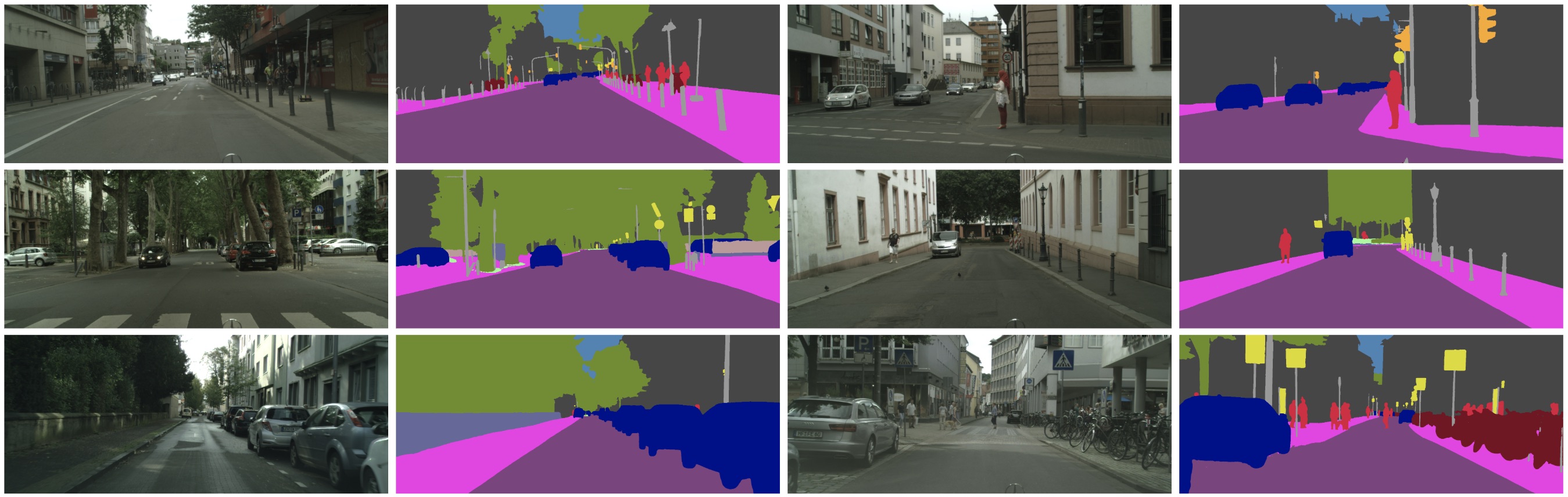} 
\vspace{-6.5mm}
\caption{Qualitative results of our method on the Cityscapes {\bf test set}. Figure shows the predicted segmentation masks.}
\label{fig:testset}
\end{figure*}

\begin{figure*}[t!]
  \vspace{-1mm}
 \centering
 \begin{footnotesize}
 \begin{tabular}{C{1.6cm}C{1.7cm}C{1.7cm}C{1.8cm}C{2cm}C{1.7cm}C{1.7cm}C{1.7cm}}
 image & ground-truth & Deeplab-v3+ & ours &  image & ground-truth & Deeplab-v3+ & ours\\[-1mm]
 \end{tabular}
  \end{footnotesize}
 \includegraphics[width=\linewidth,trim=0 0 0 0,clip]{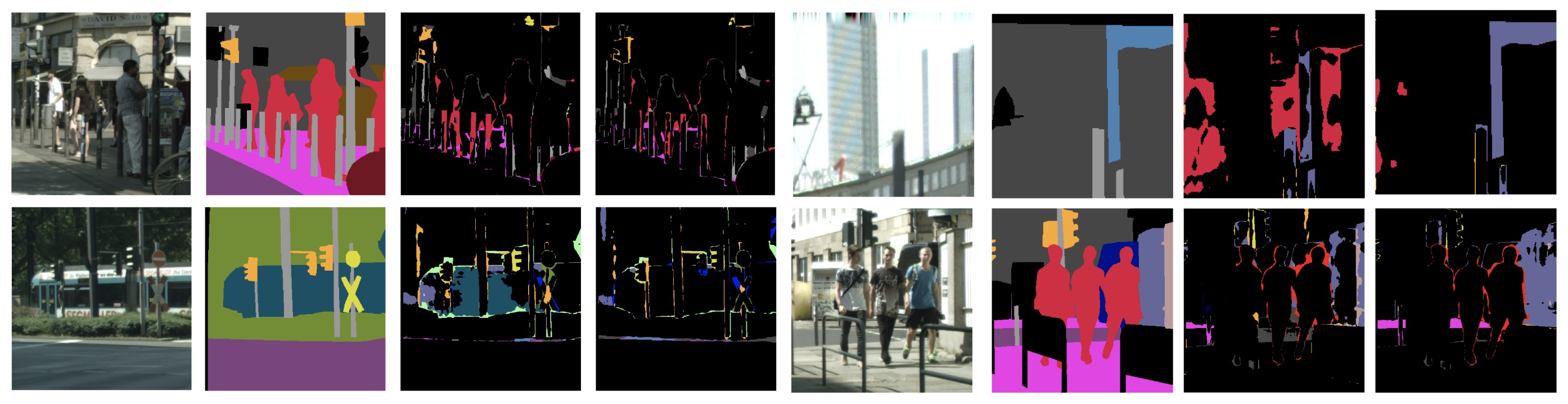} 
   \vspace{-8mm}
 \caption{Qualitative comparison in terms of {\bf errors} in predictions. Notice that our method produces more precise boundaries, particularly for smaller and thiner objects such as poles. Boundaries around people are also sharper.}
 \label{fig:errors_qualitative}
  \vspace{-1mm}
\end{figure*}

\begin{table*}[t!]
\centering 
\addtolength{\tabcolsep}{-2pt}
\resizebox{\textwidth}{!}{
  \begin{tabular}{l|c| c|c|c|c|c|c|c|c|c|c|c|c|c|c|c|c|c|c|c|c}
 Method & Coarse &  road & s.walk & build. & wall & fence & pole & t-light & t-sign & veg & terrain & sky & person & rider & car & truck & bus & train & motor & bike & mean \\
\hline \hline
        PSP-Net \cite{pspnet} & \checkmark &98.7 & 86.9 &  93.5 & 58.4 & 63.7 &  67.7 & 76.1 & 80.5 & 93.6 & 72.2  & 95.3  & 86.8 & 71.9 & 96.2 & 77.7 & 91.5 & 83.6 & 70.8 & 77.5 & 81.2 \\ 
         DeepLabV3 \cite{chen2017rethinking} & \checkmark& 98.6 & 86.2 & 93.5 & 55.2 & 63.2 & 70.0& 77.1 & 81.3 & 93.8    & 72.3&   95.9  &  87.6  & 73.4& 96.3 & 75.1 &  90.4  & 85.1 & 72.1 & 78.3 & 81.3 \\
         DeepLabV3+ \cite{deeplabv3plus2018} & \checkmark &98.7 & 87.0  & 93.9 & 59.5 & 63.7 & 71.4 &78.2 & 82.2& 94.0& 73.0 & 95.8&88.0&  73.3 & 96.4  &  78.0 & 90.9 & 83.9 & 73.8 & 78.9  & 81.9 \\
         AutoDeepLab-L \cite{liu2019auto} & \checkmark & 98.8 & 87.6  & 93.8  & 61.4  & 64.4  & 71.2 & 77.6 & 80.9 & 94.1 & 72.7 & 96.0 & 87.8 &  72.8  & 96.5  & 78.2 & 90.9  & 88.4  & 69.0  &  77.6  &  82.1 \\
         DPC \cite{chen2018searching} & \checkmark & 98.7 & 87.1 & 93.8 & 57.7 & 63.5 & 71.0 & 78.0 & 82.1 &94.0 & 73.3 & 95.4 & 88.2 &  \bf{74.5}  & \bf{96.5}  &  \bf{81.2} & \bf{93.3}  & \bf{89.0}  & \bf{74.1}  & 79.0   &   82.7 \\
   \hline \hline      
        AAF-PSP \cite{ke2018adaptive} & & 98.5   & 85.6 & 93.0 & 53.8 & 59.0  & 65.9  & 75.0   & 78.4  & 93.7  & 72.4   &  95.6 &  86.4 &70.5  & 95.9  & 73.9 & 82.7  & 76.9  &  68.7 & 76.4    &  79.1 \\
        TKCN  \cite{wu2018tree} &   &  98.4&  85.8 & 93.0 & 51.7  & 61.7  & 67.6   & 75.8  & 80.0  & 93.6   & 72.7  & 95.4  & 86.9 & 70.9  &95.9  & 64.5  & 86.9  & 81.8  & 79.6  &  77.6 &  79.5 \\ 
        \hline

\hline \hline

   \textbf{Ours (GSCNN)}& & \bf{98.7} & \bf{87.4} & \bf{94.2} & \bf{61.9} & \bf{64.6} & \bf{72.9} & \bf{79.6} & \bf{82.5} & \bf{94.3} & \bf{74.3}& \bf{96.2} & \bf{88.3} & 74.2 & 96.0 & 77.2& 90.1 & 87.7& 72.6& \bf{79.4} & \bf{82.8}\\
\hline \hline
\end{tabular}
}
\vspace{-3mm}
\caption{ 
Comparison vs  state-of-the-art methods (with/without coarse training) on the Cityscapes test set.  We only include published methods.
} 
\label{cityscapes_sota_test_set}
\end{table*}

In Fig~\ref{fig:crops}, we show the performance of our method vs baseline following the proposed distance-based evaluation method. 
Here, we find  that GSCNN performs increasingly better compared to DeeplabV3+ as the crop factor increases. The gap in performance between GSCNN and DeeplabV3+ increases from 2\% at crop factor 0 (i.e. no cropping) to close to 6\% at crop factor 400. 
This confirms that our network achieves significant improvements for smaller objects located further away from the camera.

\vspace{-4mm}
\paragraph{Cityscapes Benchmark.} 
To get optimal performance on the test set, we use our best model (\ie, regular stream is WideResNet). Training is done on an NVIDIA DGX Station using 8 GPUs with a total batch size of 16. We train this network with GCL and dual task loss for 175 epochs with a learning rate of 1e-2 with a polynomial decay policy. We also use a uniform sampling scheme to retrieve a $800 \times 800$ crop that uniformly samples from all classes.
Additionally, we use a multi-scale inference scheme using scales 0.5, 1.0 and 2.0.
We {\bf do not use coarse data} during training, due to our boundary loss which requires fine boundary annotation.

In Table~\ref{cityscapes_sota_test_set}, we compare against published state-of-the-art methods on the Cityscapes benchmark, evaluated on the test set. 
It is important to stress that  our model is not trained on coarse data.
Impressively, we can see that our model consistently outperforms very strong baselines, some of which also use extra coarse training data. At the time of this writing, our approach is also ranked as first among the published methods that do not use coarse data.

\begin{figure*}[th!]
 
\centering
\includegraphics[width=1\linewidth,trim=0 0 0 0,clip]{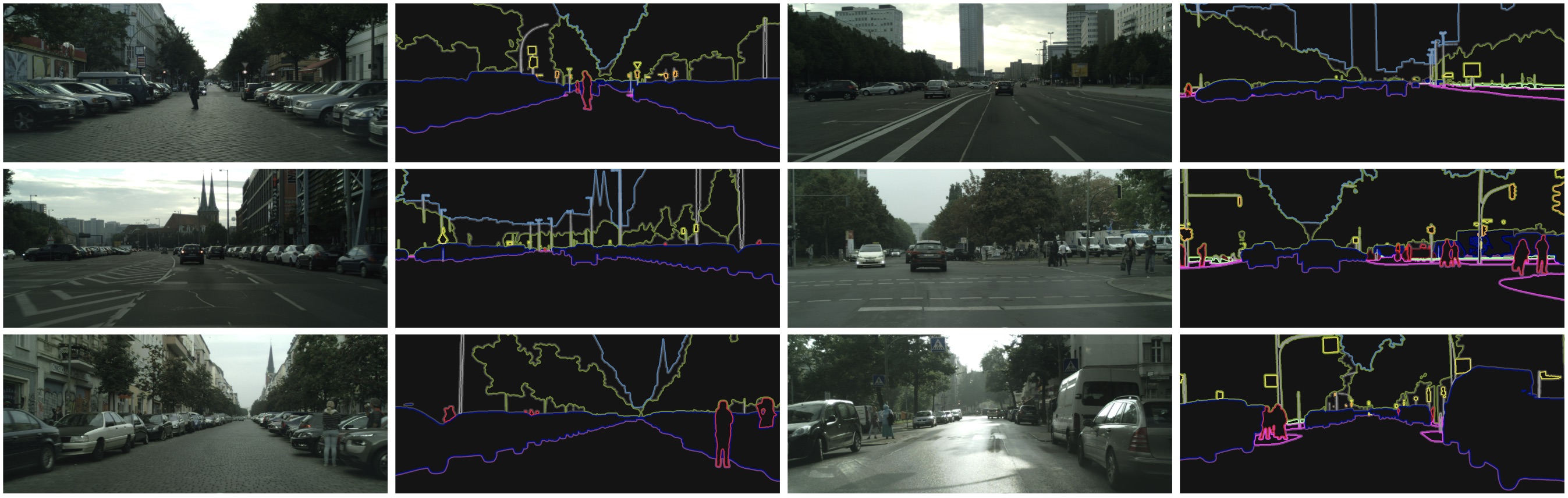} 
\vspace{-7mm}
\caption{Qualitative results on the Cityscapes test set showing the high-quality boundaries of our predicted segmentation masks. Boundaries are obtained by finding the edges of the predicted segmentation masks.}
\label{fig:testsetboundary}
 \vspace{-3mm}
\end{figure*}

\begin{figure}[t!]
 
\centering
\includegraphics[width=\linewidth,trim=0 0 0 0,clip]{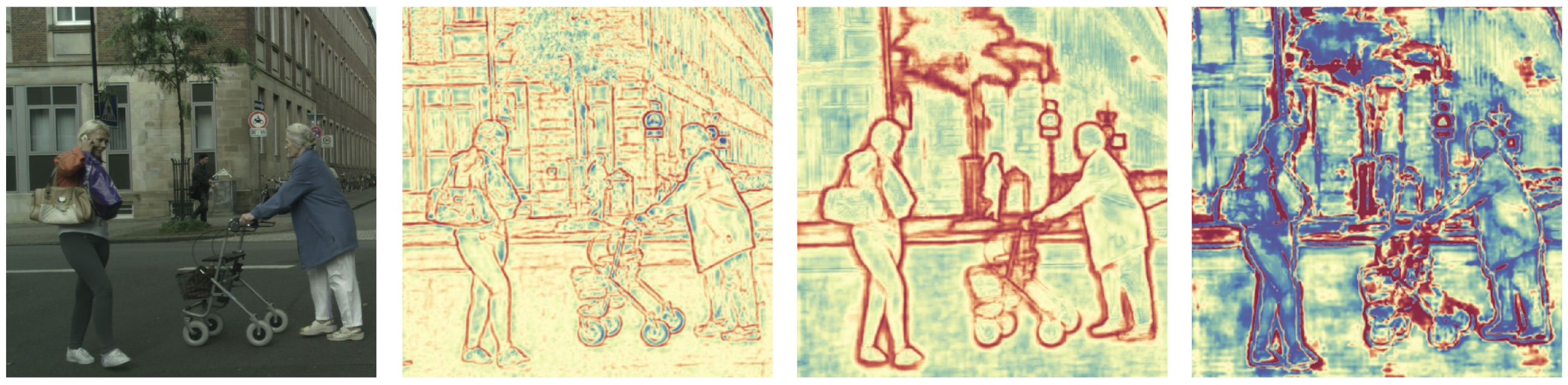}\\ 
\includegraphics[width=\linewidth,trim=0 0 0 0,clip]{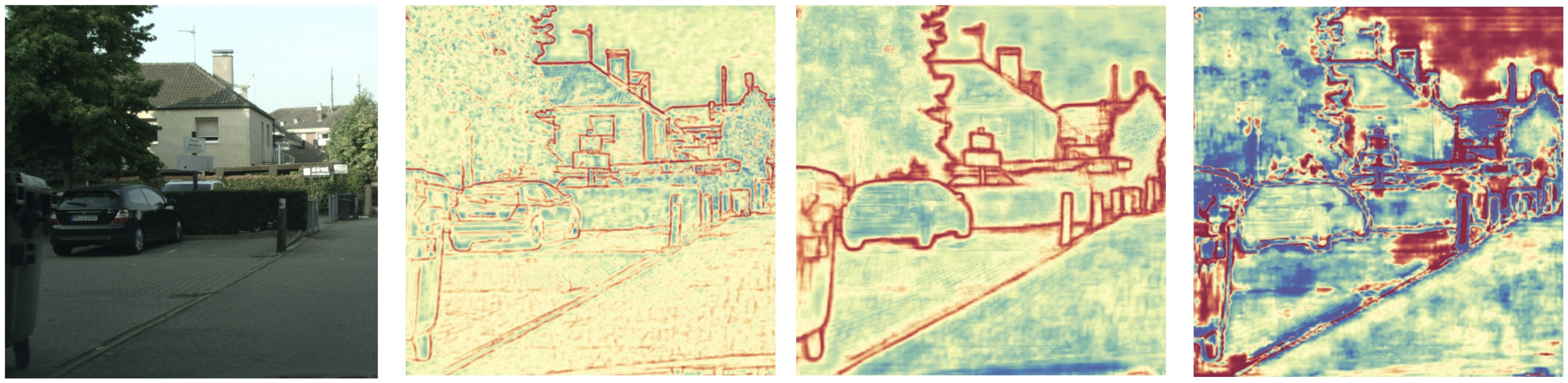}\\
\includegraphics[width=\linewidth,trim=0 0 0 0,clip]{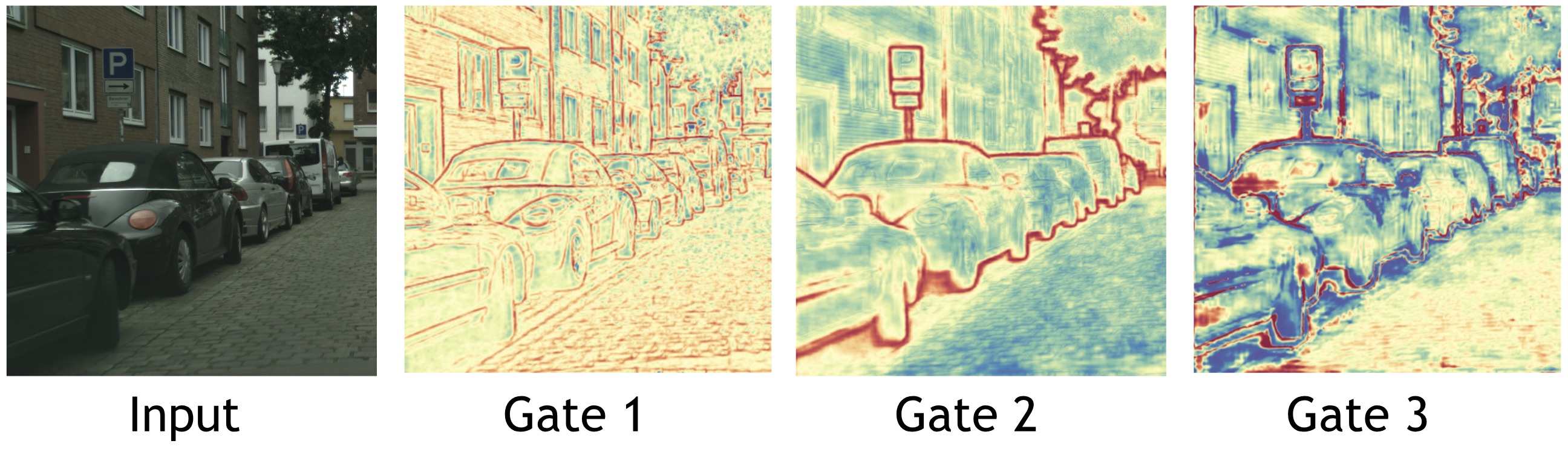}
\vspace{-6mm}
\caption{Visualization of the alpha channels from the GCLs.}
\label{fig:gates}
\vspace{-3mm}
\end{figure}

\subsection{Ablation}
In Table~\ref{tbl_effect_gcl_hf_gcl}, we evaluate the effectiveness of each component of our method using different encoder networks for the regular stream. For fairness, comparison in this table is performed with respect to our own implementation of the baseline (i.e DeepLabV3+ with different backbone architectures), trained from scratch using the same set of hyper-parameters and ImageNet initialization. Specifically, we use ResNet-50, ResNet-101 and Wide-ResNet for the backbone architectures. Here, GCL denotes a network trained with the shape stream with dual task loss, and Gradients denotes the network that also adds image gradients before the fusion layer. In our network, we use a Canny edge detector to retrieve such gradients. We see from the table that we achieve between  1 to 2 \% improvement in performance in terms of mIoU, and around  3 \% 
 for boundary alignment.

Table~\ref{tbl_effect_dual}, on the other hand, showcases the effect of the Dual Task loss in terms of  F-score for boundary alignment. We illustrate its effect at three different thresholds. Here, GCL denotes the network with the GCL shape stream trained without Dual Task Loss.
With respect to the baseline, we can observe that 
the dual loss significantly improves the performance of the model in terms of boundary accuracy.
Concretely, by adding the Dual-Task loss, we see up to 3\% improvement at the strictest regime. 

\subsection{Qualitative Results}
In Figure~\ref{fig:testset}, we provide qualitative results of our method on the Cityscapes test set.
We compare our method to the baseline by highlighting typical cases where our methods excels in Figure~\ref{fig:errors_qualitative}. Specifically, we visualize the prediction \emph{errors} for both methods.
In these zoomed images, we can see a group of people standing around an area densely populated by poles. 
Here, Deeplab v3+ fails to capture the poles and naively classifies them as humans.
Conversely, we can see that in our model poles are properly classified, 
and the error boundaries for pedestrians also thin out.
Additionally, objects such as traffic lights, which are typically predicted as an over compromising blob in Deeplab v3+ (especially at higher distances) retain their shape and structure in the output of our model. 

Fig~\ref{fig:gates} provides a visualization of the alpha channels from the GCL.
We can notice how the gates help to emphasize the difference between the boundary/region areas in the incoming feature map.
For example, the first gate emphasized very low-level edges while the second and third focus on object-level boundaries. 
As the result of gating, we obtain a final boundary map in the shape stream which accurately outlines objects and stuff classes.
This stream learns to produce high quality class-agnostic boundaries which are then fed to the fusion module. Qualitative results of the output of the shape stream are shown in Fig~\ref{fig:boundary}. 

In Figure~\ref{fig:testsetboundary}, on the other hand, we show the boundaries obtained from the final segmentation masks. Notice their accuracy on the thinner and smaller objects.

%% file: conc.tex
\vspace{-2mm}
\section{Conclusion}
\label{sec:conc}
\vspace{-1mm}
In this paper, we proposed Gated-SCNN (GSCNN), a new two-stream CNN architecture for semantic segmentation that wires shape into a separate parallel stream. 
We used a  new gating mechanism to connect the intermediate layers and a new loss function that exploits the duality between the tasks of semantic segmentation and semantic boundary prediction.
Our experiments show that this leads to a highly effective architecture that produces sharper predictions around object boundaries and significantly boosts performance on thinner and smaller objects. 
Our architecture achieves state-of-the-art results on the challenging Cityscapes dataset,  significantly improving over strong baselines.